# Adaptive Latent Factor Analysis via Generalized Momentum-Incorporated Particle Swarm Optimization

Jiufang Chen, Ye Yuan

*Abstract*—Stochastic gradient descent (SGD) algorithm is an effective learning strategy to build a latent factor analysis (LFA) model on a high-dimensional and incomplete (HDI) matrix. A particle swarm optimization (PSO) algorithm is commonly adopted to make an SGD-based LFA model's hyper-parameters, i.e, learning rate and regularization coefficient, self-adaptation. However, a standard PSO algorithm may suffer from accuracy loss caused by premature convergence. To address this issue, this paper incorporates more historical information into each particle's evolutionary process for avoiding premature convergence following the principle of a generalized-momentum (GM) method, thereby innovatively achieving a novel GM-incorporated PSO (GM-PSO). With it, a GM-PSO-based LFA (GMPL) model is further achieved to implement efficient self-adaptation of hyper-parameters. The experimental results on three HDI matrices demonstrate that the GMPL model achieves a higher prediction accuracy for missing data estimation in industrial applications.

*Index Terms*—High-Dimensional and Incomplete Data, Latent Factor Analysis, Particle Swarm Optimization, Generalized Momentum, Adaptive Model.

## I. Introduction

WITH the rapid expansion of big-data-related applications, information resources composed of multiple entities are growing in quantity as well, e.g., customers and items in a recommender system (RS) [1, 2, 33-35], and quality-of-service (QoS) in web service [3, 4, 36]. As the number of involved entities increases, the overall interaction mapping becomes difficult to observe.

In general, a high-dimensional and incomplete (HDI) matrix [3-6, 37-40] is commonly used to describe the high incomplete interactive mapping relationships. Note that in an HDI matrix, most entries that describe the unobserved data are unknown rather than "zero", while a few entries describe the observed portion [3-6], which means that the matrix is extremely sparse. Even so, it still contains wealthy and valuable content. Among the various knowledge representation models, the latent factor analysis (LFA) model is increasingly popular to extract valuable knowledge of the HDI data due to its high efficiency and scalability [7, 8]. It characterizes both users and items into the same latent factor space based on the known entries in the target HDI matrix and then estimates the missing ratings based on the corresponding vectors. Until now, many sophisticated LFA-based models are proposed to tackle the large-scale HDI data [9-11, 41]. These models are different in training schemes and model organizations, but the elemental principle is consistent.

An SGD algorithm is one of the most commonly utilized learning schemes to handle the LFA-based model [12, 42]. However, the SGD-based LFA model largely depends on the selection of its hyper-parameters, i.e., the learning rate and regularization coefficient. For instance, when the learning rate is too large or too small, the model convergence cannot reach the global optimum or suffers overshooting or even divergence [13], and when the regularization coefficient is too small or too large, the model suffers overfitting or under-fitting [14]. Therefore, both hyper-parameters need to be carefully selected. Hence, it is essential to explore a feasible and operable approach to select appropriate parameters.

Considering the commonly used hyper-parameter tuning approaches, such as manual and grid search methods, are too cumbersome and time-consuming to select an appropriate value. Some studies begin to emphasize the adaptive adjustment of hyper-parameters. For instance, serval studies [13-17] are exploring the adaptive tuning of the learning rate or regularization coefficient. Luo *et al*. [15] extend the RMSprop, AdaDelta, and Adam algorithms to the SGD algorithms to implement the learning rate adaption of the LFA-based model. Luo *et. al* [13] achieve a novel position-transitional particle swarm optimization (PSO) algorithm into an SGD-based LFA model to efficiently and automatically search for the optimal learning rate. Rendle [16] presents two optimization criteria, one is to optimize the desired LFs on the training set and the other is the regularization coefficient on the validation set. Chen *et al.* [17] propose a fine-grained adaptive regularizer that learns the regularization coefficients through training the desired LFs on the validation set, which facilitates regularization at any granularity. Meanwhile, Chen *et. al* [14] apply the standard PSO and its variant algorithm to accomplish the automatic optimization mechanism of the hyper-parametric regularization coefficient and learning rate of an SGD-based LFA model. Although it achieves a more steadily searching ability by constraining its particle position and velocity with finely-grained boundaries, its prediction accuracy and convergence rate still have an enclosure for research.

✧ J. F. Chen is with the School of Computer Science and Technology, Chongqing University of Posts and Telecommunications, Chongqing 400065, China, also with the Chongqing Institute of Green and Intelligent Technology, Chinese Academy of Sciences, and also with the Chongqing School, University of Chinese Academy of Sciences, Chongqing 400714, China (e-mail: chenjurl@outlook.com).
✧ Y. Yuan is with the College of Computer and Information Science, Southwest University, Chongqing 400715, China (e-mail: yuanyekl@gmail.com).

To address this critical issue, this paper proposes a generalized-momentum incorporated PSO (GM-PSO) algorithm that efficiently addresses the premature issues in a PSO algorithm. It calculates the momentum and "gradient" of particle velocity and position and represents their "gradient" by updating the incremental form, which is compatible with the generalized momentum method, thereby constructing the GM-PSO algorithm. The main contributions of this paper include:

**a) A GM-PSO algorithm.** It considers the historical gradient information of particles in the evolution process.
**b) A hyper-parameter-adaptive GMPL model.** It completes the hyper-parameter-adaptive training process of the learning rate and regularization coefficient.

The rest of this paper organizes as follows. Section II reviews preliminaries. Section III outlines our method in this study. Section IV discusses experiments and empirical results. Section VI summarizes the paper.

## II. PRELIMINARIES

### A. Problem Statement

An HDI matrix describes the interaction between two entity sets, i.e., a user set $U$ and an item set $I$, which are involved in relevant applications [1-4, 12-15]:

**Definition 1.** A rating matrix $R^{|U|\times|I|}$ with each entry $r_{u,i}$ quantifies some interactions between entities $u\in U$ and $i\in I$. Let K and $\Lambda$, denote $R$'s known and unknown entity sets, respectively, $R$ is an HDI matrix if $|K|\ll|\Lambda|$.

An LFA model aims to build its low-rank approximation based on K only [1-2, 12-15]:

**Definition 2.** Given $R$, $U$, $I$, $f$, an LFA model learns LF matrices $X^{|U|\times f}$ and $Y^{|I|\times f}$ to build its rank-$f$ approximation $\hat{R}=XY^T$, where $f\ll min\{|U|, |I|\}$. Note that $f$ denotes the dimension of LF space.

With it, an objective function of an LFA model can be modeled by the Euclidean distance as the distance metric:

$$\varepsilon = \sum_{r_{u,i}\in K}\left(\left(r_{u,i}-\langle x_{u,\cdot}, y_{i,\cdot}\rangle\right)^2 + \lambda\left(\|x_{u,\cdot}\|_2^2 + \|y_{i,\cdot}\|_2^2\right)\right)^2, \quad (1)$$

where $r_{u,i}$ denotes the involved entries in $R$, $x_{u,\cdot}$ and $y_{i,\cdot}$ denote the $u$-th and $i$-th row LF vectors of $X$ and $Y$, $\langle\cdot,\cdot\rangle$ computes the inner product of two entities to obtain the prediction rating, $\|\cdot\|_2$ computes the $L_2$ norm of an enclosed vector, and $\lambda$ is the regularization coefficient to avoid overfitting [18].

### B. An SGD-based LFA model

SGD is a highly effective strategy frequently adopted to solve the objective function in an LFA model [7, 10]. With it, $X$ and $Y$ in the LFs involved in (1) are updated as follows:

$$\arg\min_{X,Y}\varepsilon \overset{SGD}{\Rightarrow} \forall r_{u,i}\in K: \begin{cases} x_{u,\cdot}^{\tau}\leftarrow x_{u,\cdot}^{\tau-1}-\eta\cdot\nabla\varepsilon_{u,i}\left(x_{u,\cdot}^{\tau-1}\right), \\ y_{i,\cdot}^{\upsilon}\leftarrow y_{i,\cdot}^{\upsilon-1}-\eta\cdot\nabla\varepsilon_{u,i}\left(y_{i,\cdot}^{\upsilon-1}\right), \end{cases} \quad (2)$$

note in (2), $\tau$ and $\upsilon$ denote the current update points for $x_{u,\cdot}$ and $y_{i,\cdot}$, ($\tau$-1) and ($\upsilon$-1) denote their last update points, $\eta$ denotes the learning rate, $\varepsilon_{u,i} = \left(r_{u,i}-\langle x_{u,\cdot}^{\tau-1}, y_{i,\cdot}^{\upsilon-1}\rangle\right)^2 + \lambda\left(\|x_{u,\cdot}^{\tau-1}\|^2+\|y_{i,\cdot}^{\upsilon-1}\|^2\right)$ denotes the instant loss on $r_{u,i}\in K$ and initial states of $x_{u,\cdot}$, respectively.

Note that $x_{u,\cdot}$ and $y_{i,\cdot}$ with different symbols because of K($u$)≠K($i$). Let $e_{u,i}=r_{u,i}-\langle x_{u,\cdot}^{\tau-1}, y_{i,\cdot}^{\upsilon-1}\rangle$, then the SGD-based training rules for an LFA model applying (2) to (1) are formulated as

$$\arg\min_{X,Y}\varepsilon(X,Y)\overset{SGD}{\Rightarrow}\forall r_{u,i}\in K: \begin{cases} x_{u,\cdot}^{\tau}\leftarrow x_{u,\cdot}^{\tau-1}+\eta\cdot\left(e_{u,i}\cdot y_{i,\cdot}^{\upsilon-1}-\lambda\cdot x_{u,\cdot}^{\tau-1}\right), \\ y_{i,\cdot}^{\upsilon}\leftarrow y_{i,\cdot}^{\upsilon-1}+\eta\cdot\left(e_{u,i}\cdot x_{u,\cdot}^{\tau-1}-\lambda\cdot y_{i,\cdot}^{\upsilon-1}\right). \end{cases} \quad (3)$$

From (3), an SGD-based LFA model's performance clearly depends on $\lambda$ and $\eta$, as discussed in prior research [7, 14, 18].

## III. METHODS

### A. A Standard PSO Algorithm

A standard PSO algorithm [19-21] adopts a swarm of $q$ particles flies in $D$-dimensional space to search for the desired solution, where the movement of each particle depends on its velocity and position. More specifically, the velocity $s$ and position $h$ of the $j$-th particle at the $t$-th iteration are denoted by two vectors, i.e., $s_j^t = [s_{j,1}^t, s_{j,2}^t, \ldots, s_{j,D}^t]$ and $h_j^t = [h_{j,1}^t, h_{j,2}^t, \ldots, h_{j,D}^t]$ where $1\leq j\leq q$. During the evolution process, the $j$-th particle determines its next position based on its locally best position and the swarm's globally best position [19-21], where the latter is denoted by $pbest_j=[pbest_{j,1}, pbest_{j,2}, \ldots, pbest_{j,D}]$ and the latter by $gbest=[gbest_1, gbest_2, \ldots, gbest_D]$. The evolution scheme of the $j$-th particle at the $t$-th iteration is formulated by:

$$\begin{aligned} s_j^t &= ws_j^{t-1} + c_1 r_1\left(pbest_j^{t-1}-h_j^{t-1}\right) + c_2 r_2\left(gbest^{t-1}-h_j^{t-1}\right), \\ h_j^t &= h_j^{t-1} + s_j^t, \end{aligned} \quad (4)$$

where $w$ is the non-negative inertia constant balancing the local and global searching ability, $c_1$ and $c_2$ are cognitive and social coefficients, $r_1$ and $r_2$ are two uniform random numbers in the scale of [0, 1].

B. A Momentum-PSO Algorithm

*1) Momentum Method*

The momentum method [22] is an acceleration method that has a faster convergence speed than that of the gradient descent (GD) method [23]. Given the decision parameter $\theta$ of objective $J(\theta)$, the following update rule of the momentum-incorporated GD-based algorithm are as follows:

$$v^0 = 0, v^t = \gamma v^{t-1} + \eta \nabla J(\theta^{t-1}), \theta^t = \theta^{t-1} - v^t \tag{5}$$

where $v^0$ denotes the initial state of update velocity and normally initializes to zero, $v^t$ and $v^{t-1}$ denote the update velocity at the $t$-th and ($t$-1)-th iterations, and $\gamma$ denotes the constant balancing the effects of the previous update velocity vector and current gradient, respectively.

*2) Generalized Momentum Method*

According to formula (5), the current velocity $v^t$ is determined by the previous velocity $v^{t-1}$ and the current gradient $\nabla J(\theta^{t-1})$. However, there is no explicit gradient term in formula (5), so we generalize the standard momentum method. Specifically, the current gradient is represented as the increment form of decision parameters rather, thereby making the momentum method compatible with the PSO algorithm.

Let $\theta'^t$ represents the expected state of the decision parameters updated by the adopted learning algorithm after the $t$-th iteration, then the incremental calculation formula of the gradient is as follows

$$\Delta^t = \theta'^t - \theta^{t-1} \tag{6}$$

Let formula (6) replace in (5), the generalized form of the update velocity vector in the $t$-th iteration is obtained:

$$v^t = \gamma v^{t-1} - \Delta^t = \gamma v^{t-1} - (\theta'^t - \theta^{t-1}) \tag{7}$$

The correctness of formula (7) can be verified by using the GD algorithm as the optimization algorithm, we have

$$\begin{aligned} \theta'^t &= \theta^{t-1} - \eta \nabla_\theta J(\theta^{t-1}) \Rightarrow \Delta_t = \theta'^t - \theta^{t-1} = -\eta \nabla_\theta J(\theta^{t-1}) \\ &\Rightarrow v^t = \gamma v^{t-1} - \Delta^t = \gamma v^{t-1} + \eta \nabla_\theta J(\theta^{t-1}) \end{aligned} \tag{8}$$

According to the equivalence of formulas (5) and (8), we can obtain the generalized momentum method of velocity.

$$v^0 = 0, v^t = \gamma v^{t-1} - (\theta'^t - \theta^{t-1}), \theta^t = \theta^{t-1} - v^t \tag{9}$$

Comparing formula (9) with formula (5), we find that a standard momentum method depends on the gradient. However, the generalized one depends on the update increment by the adopted algorithm.

*3) Generalized Momentum-PSO Method*

This subsection introduces the generalized momentum method combined with the PSO algorithm method. Let $s_j^{t-1}$ and $h_j^{t-1}$ represents the state of velocity and position the $j$-th particle algorithm at ($t$-1)-th iteration. $s_j'^t$ and $h_j'^t$ denotes the velocity state of the $j$-th particle after $t$ iterations with the PSO algorithm, respectively. Therefore, the update increment caused by the PSO algorithm is calculated as

$$\Delta_j^t = [s_j'^t, h_j'^t] - [s_j^{t-1}, h_j^{t-1}] \tag{10}$$

Meanwhile, according to equations (9) and (7), we calculate the update velocity of the first iteration as follows

$$v_j^1 = \gamma v_j^0 - \Delta_j^1 = -[s_j'^1, h_j'^1] + [s_j^0, h_j^0] \tag{11}$$

where $s_j^0$ and $h_j^0$ represent the initial velocity and position of the $j$-th particle. Thus, we have

$$[s_j^1, h_j^1] = [s_j^0, h_j^0] - v_j^1 = [s_j^0, h_j^0] + [s_j'^1, h_j'^1] - [s_j^0, h_j^0] = [s_j'^1, h_j'^1] \tag{12}$$

By substituting (12) into (11), we have

$$v_j^1 = -[s_j'^1, h_j'^1] + [s_j^0, h_j^0] = -[s_j^1, h_j^1] + [s_j^0, h_j^0] \tag{13}$$

Next, we explore the second update process of particle velocity and position, thereby yielding:

$$\Delta_j^2 = [s_j'^2, h_j'^2] - [s_j^1, h_j^1] \tag{14}$$

Combined with equations (9), (12), and (14), we accomplish the status of the second iteration

$$\begin{aligned} v_j^2 &= \gamma v_j^1 - \Delta_j^2 = \gamma \left(-[s_j^1, h_j^1] + [s_j^0, h_j^0]\right) - [s_j'^2, h_j'^2] + [s_j^1, h_j^1] \\ &\Rightarrow [s_j^2, h_j^2] = [s_j^1, h_j^1] - v_j^2 = [s_j'^2, h_j'^2] + \gamma \left([s_j^1, h_j^1] - [s_j^0, h_j^0]\right) \end{aligned} \tag{15}$$

we achieve the following compact form for the PSO-based evolution rule:

$$\begin{cases} t=1: \left[s_j^1, h_j^1\right] = \left[s_j'^1, h_j'^1\right] \\ t \geq 2: \left[s_j^t, h_j^t\right] = \left[s_j'^t, h_j'^t\right] + \gamma \left( \left[s_j^{t-1}, h_j^{t-1}\right] - \left[s_j^{t-2}, h_j^{t-2}\right] \right) \end{cases} \quad (16)$$

Note that with the PSO algorithm, the velocity and position of each particle are updated by the generalized momentum method. Therefore, by combining equations (4) and (16), the following update rules can be obtained:

$$\begin{cases} t=1: \begin{cases} s_j^1 = s_j^0 + c_1 r_1 \left(pbest_j^0 - h_j^0\right) + c_2 r_2 \left(gbest^0 - h_j^0\right) \\ h_j^1 = h_j^0 + s_j^1 \end{cases} \\ t \geq 2: \begin{cases} s_j^t = s_j^{t-1} + c_1 r_1 \left(pbest_j^{t-1} - h_j^{t-1}\right) + c_2 r_2 \left(gbest^{t-1} - h_j^{t-1}\right) + \gamma \left(s_j^{t-1} - s_j^{t-2}\right) \\ h_j^t = h_j^{t-1} + s_j^t + \gamma \left(h_j^{t-1} - h_j^{t-2}\right) \end{cases} \end{cases} \quad (17)$$

Note that the momentum coefficient $\gamma$ is introduced as a new hyper-parameter, we automatically increase it in the range of (0.4, 1.4) according to [24]. It can increase the influence of momentum and accelerate the convergence [25], on the other hand, it can help particles jump out of the local saddle point. The implementation is as follows

$$\gamma \leftarrow \min\left(\gamma_{\min} + 0.1 \left\lfloor \frac{t}{m} \right\rfloor, \gamma_{\max}\right) \quad (18)$$

where $\gamma_{\max}=1.4$, $\gamma_{\min}=0.4$, $t$ represents the current iteration count, $\lfloor \cdot \rfloor$ represents the momentum coefficient varies every $m$ iteration counts, where $m$ indicates that the fixed number is set to 5.

*C. A GMPL Model*

In a GM-PSO algorithm, the $j$-th particle maintains an individual group of hyper-parameters, i.e., $\eta_j$ and $\lambda_j$. Thus, its position during the evolution is given as:

$$h_j = \left[\eta_j, \lambda_j\right] \quad (19)$$

On the other hand, all particles are incorporated with the same pair of LF matrices $X$ and $Y$. Thus, each evolution iteration consists of $q$ sub-iterations, where at the $j$-th sub iteration of the $t$-th evolution iteration, $X$ and $Y$ are updated as follows:

$$\arg\min_{X,Y} \varepsilon(X,Y) \stackrel{GM-PSO-SGD}{\Rightarrow} \forall r_{u,i} \in K: \Rightarrow \begin{cases} x_{(j)u,\cdot}^{\tau} \leftarrow x_{(j)u,\cdot}^{\tau-1} + \eta_j^t \cdot \left(e_{(j)u,i} \cdot y_{(j)i,\cdot}^{\upsilon-1} - \lambda_j^t \cdot x_{(j)u,\cdot}^{\tau-1}\right), \\ y_{(j)i,\cdot}^{\upsilon} \leftarrow y_{(j)i,\cdot}^{\upsilon-1} + \eta_j^t \cdot \left(e_{(j)u,i} \cdot x_{(j)u,\cdot}^{\tau-1} - \lambda_j^t \cdot y_{(j)i,\cdot}^{\upsilon-1}\right), \end{cases} \quad (20)$$

where the subscript ($j$) on $x_{u,\cdot}$ and $y_{i,\cdot}$ denotes that their current update is linked with the $j$-th particle, i.e., $\eta_j$ and $\lambda_j$. Note that restricting the position and velocity of particles to ensure that each particle can fly within a predetermined boundary, thereby yielding:

$$h_j^t \stackrel{Bounded}{=} \begin{cases} \eta_j^t = \max\left(\eta_{\min}, \min\left(\eta_{\max}, \eta_j^t\right)\right), \\ \lambda_j^t = \max\left(\lambda_{\min}, \min\left(\lambda_{\max}, \eta_j^t\right)\right), \end{cases} \quad (21a)$$

$$s_j^t \stackrel{Bounded}{=} \max\left(s_{\min}, \min\left(s_{\max}, s_j^t\right)\right) \quad (21b)$$

where the searching intervals $[\eta_{\min}, \eta_{\max}]$ and $[\lambda_{\min}, \lambda_{\max}]$ are set as $[2^{-13}, 2^{-7}]$ and $[2^{-7}, 2^{-1}]$. Considering the fitness function, we set it according to the contribution of active particles to reducing the loss of the whole swarm.

$$A_0^t = A_q^{t-1}, F_j^t = \frac{A_j^t - A_{j-1}^t}{A_q^t - A_q^{t-1}}, \quad (22)$$

where $F_j^t$ denotes the contribution of the $j$-th particle at $t$-th iteration, and $A_j^t$ represents the minimizing prediction error of the $j$-th particle at $t$-th iteration, respectively. We formulate the quantizing function in the form of root mean squared error (RMSE) as:

$$A_j = \sqrt{\left[\sum_{r_{u,i} \in \Omega} \left(r_{(j)u,i} - \hat{r}_{(j)u,i}\right)^2\right] / |\Omega|}, \quad (23)$$

where $\Omega$ denotes the validation set and is disjoint with the training set K and testing set $\Gamma$, $|\cdot|_{abs}$ calculates the absolute value of an enclosed number, and $\hat{r}_{(j)u,i}$ denotes the prediction generated by the LFs achieved with the hyper-parameter settings following the $j$-th particle, respectively.

On the other hand, the improvement rate ($Ir$) of a particle $j$ in GM-PSO is defined according to [26]:

$$Ir\left(h_j^t\right) = \frac{F_j^{t-1} - F_j^t}{e^{\left|h_j^{t-1} - h_j^t\right|}}, t \geq 2 \quad (24)$$

where $F_j^{t-1}$ is the contribution of $j$-th particle $h_j^{t-1}$ on ($t$-1)-th iteration, $|h_j^{t-1}-h_j^t|$ denotes the Euclidean distance between $h_j^{t-1}$ and $h_j^t$ as setting according to [26, 27]. From Eq. (24), we can find that a higher improvement rate ($Ir$) means a particle $j$ obtains a larger contribution at the smaller distance of two iterations.

Then, we design the update rules for *pbest$_j$* and *gbest*, which is as follows:

$$\begin{cases} \begin{cases} pbest_j^t = \begin{cases} pbest_j^{t-1}, & F_j^t \leq F_j^{t-1}, \\ h_j^t, & F_j^t > F_j^{t-1}; \end{cases} \\ gbest^t = \begin{cases} gbest^{t-1}, & F_j^t \leq F_{j-1}^t, \\ h_j^t, & F_j^t > F_{j-1}^t. \end{cases} \end{cases} t=1 \\ \begin{cases} pbest_j^t = \begin{cases} pbest_j^{t-1}, & F_j^t \leq F_j^{t-1}, \\ h_j^t, & F_j^t > F_j^{t-1}; \end{cases} \\ gbest^t = \begin{cases} gbest^{t-1}, & Ir_j^t \leq Ir_j^{t-1}, \\ h_j^t, & Ir_j^t > Ir_j^{t-1}. \end{cases} \end{cases} t \geq 2 \end{cases} \quad (25)$$

Based on (19)-(25), we achieve the GMPL model.

*D. Algorithm Design and Analysis*

Based on previous sections, we design the Algorithm GMPL. As shown in Algorithm GMPL, we summarize its computational cost as follows:

$$T_{GMPL} = \Theta\left(n \times q \times (|K|+|\Omega|) \times f\right). \quad (26)$$

Note that (26) holds the condition $(|K|+|\Omega|) \times f \gg \max\{|U|, |I|\}$, which is constantly satisfied in industrial applications. It can be seen that the computational cost of the proposed method is linear with $(|K|+|\Omega|)$, and it can extend to large-scale datasets.

As shown in Algorithm GMPL, we adopt several auxiliary matrices to store relevant data: 1) caching the LF matrices $X$ and $Y$, whose storage costs sum up to $\Theta((|U|+|I|) \times f)$; 2) caching the auxiliary matrices of particles to complete the evolution of the hyper-parameter, i. e. *S, H, pbest, F* and *Ir*. Since those auxiliary matrices of particles are far less than $\min\{|U|, |I|\}$, such a storage burden is easy to resolve. Algorithm GMPL's storage cost is given as:

$$S_{GMPL} = (|U|+|I|) \times f, \quad (27)$$

which is linear with the involved entity count in $R$.

| Algorithm GMPL | |
|---|---|
| **Operation** | **Cost** |
| **while not** converge **and** $t \leq N$ **do** | $\times n$ |
|   **for** $j$=1 **to** $q$ | $\times q$ |
|     Make $H_j$ and $S_j$ evolve via (17). | $\Theta(1)$ |
|     Bound $H_j$ and $S_j$ via (21). | $\Theta(1)$ |
|   Update $\gamma$ via (18). | $\Theta(1)$ |
|   **for** $j$=1 **to** $q$ | $\times q$ |
|     **for each** $r_{u,i}$ in K | $\times |K|$ |
|       Update $x_{u,\cdot}$ and $y_{i,\cdot}$ via (10). | $\Theta(f)$ |
|     Compute $A_j^t$ via (23). | $\Theta(|\Omega|*f)$ |
|   **for** $j$=1 **to** $q$ | $\times q$ |
|     Compute $F_j^t$ with (22) | $\Theta(1)$ |
|     Compute $Ir_j^t$ with (24) | $\Theta(1)$ |
|     Update *pbest$_j$* and *gbest* with (25) | $\Theta(1)$ |
|   $t=t+1$ | $\Theta(1)$ |
| **Output:** $X, Y$ | |

IV. EXPERIMENTAL RESULTS AND ANALYSIS

*A. General Settings*

**Evaluation Protocol.** For industrial applications, estimating the missing data of an HDI matrix is the main motivation expected for discovering the whole interactions between involved known sparse entities. Hence, we adopt the most used RMSE as the evaluation metrics [1, 3-7].

$$RMSE = \sqrt{\left(\sum_{r_{u,i} \in \Gamma}(r_{u,i}-\hat{r}_{u,i})^2\right) \Big/ |\Gamma|},$$

where $\hat{r}_{u,i}$ denotes the generated prediction for the testing instance $r_{u,i} \in \Gamma$, $|\cdot|$ represents the cardinality of a given set, $|\Gamma|$ denotes the size of the testing dataset $\Gamma$, respectively.

**Datasets.** In our experiments, three HDI matrices collected from industry applications are adopted, as detailed in Table I.

TABLE I. DETAILS OF EXPERIMENTAL DATASETS

| No. | Name | Row | Column | Known Entries | Density |
|---|---|---|---|---|---|
| D1 | Jester [28] | 24983 | 100 | 1,186,324 | 72.41% |
| D2 | ML10M [29] | 71,567 | 65,133 | 10,000,054 | 1.31% |
| D3 | Flixster [30] | 147,612 | 48,794 | 8,196,077 | 0.11% |

Note that each dataset is randomly split into ten disjoint subsets for implementing the ratio of 70%-10%-20% as the train-validation-test settings to achieve objective results. More specifically, a trained model is built on seven subsets $K$, a validated model is built on one subset $\Omega$, and a tested model is built on two subsets $\Gamma$ to verify the performance of its outcomes. The final results are based on an average of ten repetitions. The termination condition is uniform for all compared models, i.e., the iteration threshold is 1000, and the threshold of error for two consecutive iterations is $10^{-5}$.

### A. Comparison with LFA Models

**Compared Models.** In this subsection, we compared the GMPL model with several LFA models. Note that the results are conducted on a Tablet with a 2.1GHz Xeon(R) CPU and 256-GB RAM. The programming language is JAVA SE8U131. A detailed comparison is performed in Table II and the manually-tuning hyper-parameters of M1-4 are summarized in Table III.

TABLE II. DETAILS OF COMPARED MODEL

| Model | Description |
|---|---|
| M1 | An SGD-based LFA model [7]. |
| M2 | A PSO-based LFA model [14]. |
| M3 | An adaptive weighting PSO-based [31] LFA model. |
| M4 | The proposed model of this study. |

TABLE III. HYPER-PARAMETERS CALL FOR MANUALLY TUNING IN EACH MODEL.

| Model | Description |
|---|---|
| M1 | $\eta$, learning rate; $\lambda$, regularization coefficient |
| M2-M3 | $\eta$, $\lambda$ self-adaptation |
| M4 | $\eta$, $\lambda$, $\gamma$, self-adaptation |

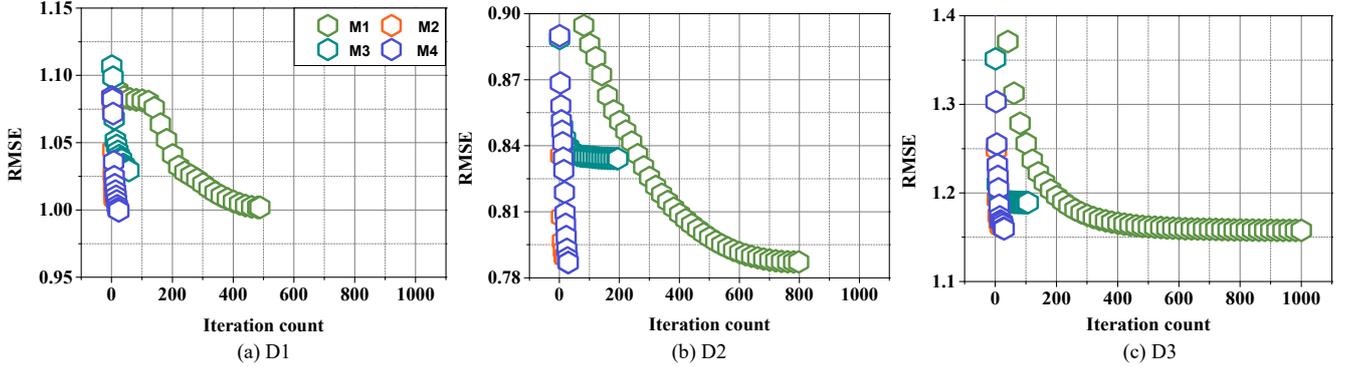

Fig. 1. M1-4's training curves on D1-3. Note that all panels share the same legend in panel (a).

TABLE IV. LOWEST RMSE AND ITS TOTAL TIME COST. A LOWER VALUE IS BETTER.

| | Case | M1 | M2 | M3 | M4 |
|---|---|---|---|---|---|
| D1 | RMSE | 1.0001$_{\pm 6E-5}$ | 1.0063$_{\pm 3E-3}$ | 1.0278$_{\pm 9E-3}$ | **0.9982$_{\pm 4E-4}$** |
| | Time-RMSE | 107.5$_{\pm 2.0}$ | **20.1$_{\pm 1.9}$** | 134.1$_{\pm 23.3}$ | 57.9$_{\pm 14.9}$ |
| D2 | RMSE | 0.7872$_{\pm 5E-4}$ | 0.7887$_{\pm 2E-3}$ | 0.7925$_{\pm 1E-5}$ | **0.7867$_{\pm 4E-4}$** |
| | Time-RMSE | 932.1$_{\pm 11.2}$ | **285.3$_{\pm 15.6}$** | 2596.7$_{\pm 54.1}$ | 307.8$_{\pm 14.5}$ |
| D3 | RMSE | 0.9445$_{\pm 2E-4}$ | 0.9477$_{\pm 7E-4}$ | 0.9861$_{\pm 4E-5}$ | **0.9409$_{\pm 9E-4}$** |
| | Time-RMSE | 1060.3$_{\pm 17.5}$ | **216.8$_{\pm 63.4}$** | 1257.4$_{\pm 31.9}$ | 318.7$_{\pm 32.4}$ |

**Experimental Settings.** To maintain the fairness of comparison, we use the following settings for all models:

a) The initialization of LF matrices is randomly generated to remove performance bias;
b) The LF dimension is uniformly set as $f$=20 to balance the computational efficiency and representative learning ability of each model [1, 3-7].
c) For M2-4, the hyper-parameters of PSO are fixed following [20], i.e., w=0.729, $c_1$=$c_2$=2, $q$=10, $r_1$ and $r_2$ are two uniformly distributed random numbers in a range of [0, 1], the swarm size $q$=10 according to [13, 14], respectively. Note the inertial weight in M3 is adaptive.

**Results and Analysis.** Tables IV summarizes M1-4's lowest RMSE, and corresponding total time costs, respectively. Fig. 1 depicts their training curves in RMSE. From these results, we have the following findings:

a) **The GMPL model implements effective hyper-parameters adaptation due to the GM-PSO algorithm.** As shown in Table IV, for instance, on D1, M4's RMSE is 0.9982, which is 0.19% lower than 1.0001 by M1, 0.80% lower than 1.0063 by M5, 2.88% lower than 1.0278 by M7, respectively, which shows a GMPL model generally outperforms to other models in terms of prediction accuracy.
b) **GMPL has a competitive advantage in computational efficiency when compared with its peers.** According to the analysis in Section III (D), the per iteration cost of the GMPL model is about $q$ (i.e. $q=10$) times that of the standard LFA-based model, but its total time cost is relatively low due to its fast convergence. However, the time efficiency of standard PSO is higher than that of GM-PSO, which is at the expense of prediction accuracy. Meanwhile, compared with the SGD-based LFA model and advanced PSO-based LFA model, the computational efficiency of GMPL owns advantageous.
c) **Summary.** Based on the experimental results and corresponding analyses, we summarize that: a) GMPL significantly alleviates the premature convergence issue owing to its incorporation of the proposed GM-PSO algorithm, and; b) GMPL implements self-adaptation of hyper-parameters with competitive prediction accuracy for missing data of an HDI matrix along with low time cost.

## V. Conclusions

In this paper, we achieve the effective adaptation of the hyper-parametric learning rate and regularization coefficient through our proposed GM-PSO algorithm. Compared with the advanced PSO algorithms, some extended PSO algorithms will increase the time complexity of the LFA model. The GM-PSO algorithm we proposed integrates the idea of momentum algorithm into the standard PSO, realizes the adaptation of multiple hyper-parameters of the SGD-based LFA model, shares the same hyper-parametric learning mechanism to reduce time complexity and computational efficiency, and achieves better prediction accuracy for the missing value of HDI matrix. In future work, we aim to extend the GM-PSO algorithm to other models, such as the protein complex detection model [5, 43], tensor decomposition model [32, 44, 45], and so on.